\documentclass[letterpaper, 10 pt, conference]{ieeeconf}
\IEEEoverridecommandlockouts       

\usepackage[usenames,dvipsnames,svgnames]{xcolor}
\usepackage{times} 
\usepackage{amsmath} 
\usepackage{amssymb}  
\usepackage{algorithm}
\usepackage[noend]{algpseudocode}
\usepackage{comment}
\usepackage{graphicx}
\usepackage{subfigure}
\usepackage{tikz}
\usepackage{etoolbox}
\usepackage{xspace}
\usepackage[letterspace=0,tracking=true]{microtype}
\usepackage{comment}
\usepackage{hyperref}
\usepackage[utf8]{inputenc}
\usepackage[T2A,T1]{fontenc}
\usepackage{booktabs}
\usepackage{paralist}
\usepackage{cleveref}
\usepackage[export]{adjustbox}

\makeatletter
\let\NAT@parse\undefined
\makeatother
\usepackage[numbers]{natbib}

\hypersetup{%
  pdfborder = {0 0 0},
  bookmarksdepth = section,
  citecolor = DarkGreen,
  linkcolor = DarkBlue,
  urlcolor = DarkBlue,
}%

\graphicspath{{images/}}

\definecolor{Red}{rgb}{1,0,0}
\definecolor{Green}{rgb}{0,0.69,0}
\definecolor{Blue}{rgb}{0,0,1}
\definecolor{LightBlue}{rgb}{0,0.5,1}
\definecolor{veryLightBlue}{rgb}{0.85,0.98,1}
\definecolor{veryLightGreen}{rgb}{0.6,1,0.6}
\definecolor{Skin}{rgb}{1,0.71,0.69}
\definecolor{Grey}{rgb}{0.5,0.5,0.5}
\definecolor{LightGrey}{rgb}{0.6,0.6,0.6}
\definecolor{Black}{rgb}{0,0,0}
\definecolor{White}{rgb}{1,1,1}
\newcommand{\red}{\color{Red}}
\newcommand{\green}{\color{Green}}
\newcommand{\blue}{\color{Blue}}

\newcommand{\grey}{\color{Grey}}

\newcommand{\orange}{\color{Bittersweet}}

\newbool{blind}
\setbool{blind}{false}

\ifbool{blind}
{\AtBeginEnvironment{blind}{\comment}%
  \AtEndEnvironment{blind}{\endcomment}}{}

\makeatletter
\DeclareRobustCommand\onedot{\futurelet\@let@token\@onedot}
\def\@onedot{\ifx\@let@token.\else.\null\fi\xspace}

\def\etal{\emph{et al}\onedot}
\newcommand{\eg}{e.g.,\xspace}

\newcommand{\ie}{i.e.,\xspace}

\makeatother

\title{\LARGE \bf
  Deep compositional robotic planners\\ that follow natural language commands
}

\ifbool{blind}{}{%
  \author{Yen-Ling Kuo, Boris Katz, and Andrei Barbu%
    \thanks{This work was supported by the Center for Brains, Minds and
      Machines, NSF STC award 1231216, the Toyota Research
      Institute, the DARPA GAILA program, the ONR Science of
      Artificial Intelligence, and CBMM-Siemens Graduate Fellowship.}%
    \thanks{\protect\raggedright CSAIL and CBMM, MIT
      {\tt\small \{ylkuo,boris,abarbu\}@mit.edu}}%
  }}

\usetikzlibrary{calc}
\usetikzlibrary{fit}
\usetikzlibrary{positioning}
\usetikzlibrary{decorations.markings}
\usetikzlibrary{shapes.geometric}
\usetikzlibrary{shapes.multipart}
\usetikzlibrary{bayesnet}
\usetikzlibrary{tikzmark}
\usetikzlibrary{backgrounds} 
\usetikzlibrary{shapes}
\usetikzlibrary{3d}
\usetikzlibrary{arrows.meta}
\usetikzlibrary{shapes.geometric}

\definecolor{pinegreen}{cmyk}{0.92,0,0.59,0.25}
\definecolor{royalblue}{cmyk}{1,0.50,0,0}
\tikzstyle{cblue}=[circle, draw, thin,fill=cyan!20, scale=0.8]
\tikzstyle{obs}=[circle, draw, thin,fill=gray!20, scale=0.8]
\tikzstyle{qgre}=[circle, draw, scale=0.8]
\tikzstyle{rpath}=[thick, black, opacity=0.4]

\tikzstyle{pathnode}=[circle, draw, scale=0.04]
\tikzstyle{graphnode}=[circle, draw, scale=0.15,ultra thin]
\tikzstyle{graphedge}=[-{Latex[black,length=0.15ex,width=0.15ex]}, shorten >= 0.04ex, ultra thin]
\tikzstyle{graphdashed}=[dash pattern=on 0.2pt off 0.1pt]

\newcommand{\bugtrap}%
{
  \draw[ultra thin, purple] (0,0,0)--(0,1,0) (0,1,0)--(1,1,0) (1,1,0)--(1,0,0) (1,0,0)--(0,0,0);
  \coordinate (a) at (0.25,0.25,0);
  \coordinate (b) at ($(a)+(0.21,0,0)$);
  \coordinate (c) at ($(b)+(0,0.24,0)$); 
  \coordinate (d) at ($(a)+(0,0.5,0)$); 
  \coordinate (e) at ($(d)+(0.5,0,0)$); 
  \coordinate (f) at ($(e)+(0,-0.5,0)$); 
  \coordinate (g) at ($(f)+(-0.21,0,0)$); 
  \coordinate (h) at ($(g)+(0,0.24,0)$);
  \draw[ultra thin, purple] (a)--(b) (b)--(c) (d)--(a) (d)--(e) (e)--(f) (f)--(g) (g)--(h);
}

\newcommand{\plannerrnn}[4]%
{
  \node[graphnode, above=1 of start, xshift=-10ex, xshift=#2, yshift=0ex, yshift=#3, #4] (gstartX#1) {};
  \node[graphnode, above=1 of n1, xshift=-1ex, xshift=#2, yshift=0ex, yshift=#3, #4] (gn1X#1) {};
  \node[graphnode, above=1 of n2, xshift= 1ex, xshift=#2, yshift=5ex, yshift=#3, #4] (gn2X#1) {};
  \node[graphnode, above=1 of n3, xshift=-1ex, xshift=#2, yshift=-5ex, yshift=#3, #4] (gn3X#1) {};
  \node[graphnode, above=1 of n4, xshift=2ex, xshift=#2, yshift=0ex, yshift=#3, #4] (gn4X#1) {};
  \node[graphnode, above=1 of n5, xshift=8ex, xshift=#2, yshift=5ex, yshift=#3, #4] (gn5X#1) {};
  \node[graphnode, above=1 of n6, xshift=10ex, xshift=#2, yshift=-5ex, yshift=#3, #4] (gn6X#1) {};

  \draw[graphedge, graphdashed] (gstartX#1) -- (start);
  \draw[graphedge, graphdashed] (gn1X#1) -- (n1);
  \draw[graphedge, graphdashed] (gn2X#1) -- (n2);
  \draw[graphedge, graphdashed] (gn3X#1) -- (n3);
  \draw[graphedge, graphdashed] (gn4X#1) -- (n4);
  \draw[graphedge, graphdashed] (gn5X#1) -- (n5);
  \draw[graphedge, graphdashed] (gn6X#1) -- (n6);
  \draw[graphedge, graphdashed] (gn6X#1) -- (n7);
  \draw[graphedge] (gstartX#1) -- (gn1X#1);
  \draw[graphedge] (gstartX#1) -- (gn2X#1);
  \draw[graphedge] (gn1X#1) -- (gn3X#1);
  \draw[graphedge] (gn1X#1) -- (gn4X#1);
  \draw[graphedge] (gn4X#1) -- (gn5X#1);
  \draw[graphedge] (gn4X#1) -- (gn6X#1);
}

\begin{document}

\maketitle
\thispagestyle{empty}
\pagestyle{empty}


\begin{abstract}
  We demonstrate how a sampling-based robotic planner can be augmented to learn
  to understand a sequence of natural language commands in a continuous
  configuration space to move and manipulate objects.
  Our approach combines a deep network structured according to the parse of a
  complex command that includes objects, verbs, spatial relations, and
  attributes, with a sampling-based planner, RRT.
  A recurrent hierarchical deep network controls how the planner explores the
  environment, determines when a planned path is likely to achieve a goal, and
  estimates the confidence of each move to trade off exploitation and
  exploration between the network and the planner.
  Planners are designed to have near-optimal behavior when information about the
  task is missing, while networks learn to exploit observations which are
  available from the environment, making the two naturally complementary.
  Combining the two enables generalization to new maps, new kinds of obstacles,
  and more complex sentences that do not occur in the training set.
  Little data is required to train the model despite it jointly acquiring a CNN
  that extracts features from the environment as it learns the meanings of
  words.
  The model provides a level of interpretability through the use of attention
  maps allowing users to see its reasoning steps despite being an end-to-end
  model.
  This end-to-end model allows robots to learn to follow natural language
  commands in challenging continuous environments.
\end{abstract}

\section{INTRODUCTION}

When you carry out a command uttered in natural language, you combine your
knowledge about the task to be performed and how it was carried out in the past
with reasoning about the consequences of your actions.
Thinking about the task allows you to choose actions that are likely to make
progress, and it is most useful when the path forward is clearly understood in an
environment that has been experienced before.
Thinking about the consequences of your actions allows you to handle new
environments and obstacles, and it is most useful where a task must be performed in
a novel way.
Generally, prior work has excelled at one of these but not both.
Powerful models can control agents but do so from moment to moment without
planning complex
actions~\citep{Blukis:18visit-predict,misra2018mapping,shah2018follownet}.
Planners on the other hand efficiently explore configuration spaces, often by
building search trees~\citep{Hsu1997:NarrowPassage,Karaman2011:rrtstar}, but
require a target final
configuration~\citep{chen2019learning,kuo2018deep,lee2018gated} or a symbolic
specification of constraints~\citep{tellex2011understanding,paul2017temporal}.

We demonstrate an end-to-end model that both reasons about a task and plans its
action in a continuous domain resulting in a robot that can follow linguistic
commands.\footnote{We will release the source code and models upon publication.}
This is the first model to perform end-to-end navigation and manipulation tasks
given natural language commands in continuous environments without symbolic
representations.
It integrates a planner with a compositional hierarchical recurrent network.
The recurrent network learns which actions are useful toward a goal specified in
natural language while the planner provides resilience when the situation
becomes unclear, novel or too complicated.
This process frees the network from having to learn the minutia of planning and
allows it to focus on the overall goal while gaining robustness to novel
environments.

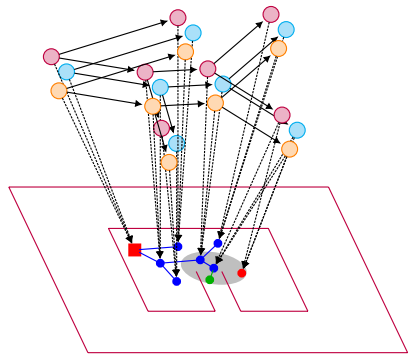
\begin{figure}
  \centering%
  \vspace{19ex}
  \scalebox{0.85}{%
  \begin{tikzpicture}[square/.style={regular polygon,regular polygon sides=4}]
    \begin{scope}[rotate around x=40,transform canvas={scale=5},yshift=-12,xshift=-10]
      \bugtrap
      \node[pathnode, square, scale=2, red, fill=red] (start) at ($(a)+(0.05,0.37,0)$) {};
      \node[pathnode, blue, fill=blue] (n1) at ($(start)+(0.06,-0.08,0)$) {};
      \node[pathnode, blue, fill=blue] (n2) at ($(start)+(0.14,0.02,0)$) {};
      \node[pathnode, blue, fill=blue] (n3) at ($(n1)+(0.024,-0.11,0)$) {};
      \node[pathnode, blue, fill=blue] (n4) at ($(n1)+(0.13,0.02,0)$) {};
      \node[pathnode, blue, fill=blue] (n5) at ($(n4)+(0.08,0.10,0)$) {};
      \node[pathnode, blue, fill=blue] (n6) at ($(n4)+(0.03,-0.05,0)$) {};
      \node[pathnode, red, fill=blue] (n7) at ($(n6)+(0.08,-0.03,0)$) {}; 
      \node[pathnode, Green, fill=blue] (n8) at ($(n6)+(-0.03,-0.07,0)$) {};
      \draw[ultra thin,blue] (start)--(n1) (start)--(n2) (n1)--(n3) (n1)--(n4) (n4)--(n5) (n4)--(n6);
      \draw[ultra thin, Green] (n6)--(n8);
      \begin{pgfonlayer}{background}
        \begin{scope}[transform canvas={scale=5}]
          \fill[opacity=0.5, fill=gray] (n6) circle (0.1);
        \end{scope}
      \end{pgfonlayer}
      \plannerrnn{1}{1ex}{0ex}{cyan,fill=cyan!30}
      \plannerrnn{2}{-1ex}{2ex}{purple,fill=purple!30}
      \plannerrnn{3}{0ex}{-2.5ex}{orange,fill=orange!30}
    \end{scope}
  \end{tikzpicture}}
  \vspace{10ex}
  \caption{We augment a sampling-based planner, RRT, with a hierarchical
    recurrent network that encodes the meaning of a natural-language command the
    robot must follow. Just as with a traditional planner, the robot mentally
    explores the space around a {\red start location} building a {\blue search
      tree} to find a good path in its configuration space. Unlike a traditional
    planner, we do not specify a goal as a location, but instead rely on a
    neural network to score how likely any position in the configuration space
    is to be an end state while considering the past history of the robot's
    actions and its observation of the environment. The structure of the RNNs
    mirrors that of the search tree, with each splitting off as different
    decisions are considered. At each time step, the RNNs {\grey observe the
      environment}, and can adjust the sampling process of the planner to avoid
    moving in undesirable locations (in this case, the tree is not expanded
    toward the red circle, and instead adjusted to go down the passageway
    through the green circle). See~\cref{fig:rnn-model} for details on the
    structured RNNs and how they encode the structure of sentences as relationships
    between recurrent models.}
  \label{fig:model-overview}
\end{figure}

To execute a command, the model proceeds as a traditional sampling-based planner
with an additional input of a natural language command;
see~\cref{fig:model-overview}.
A collection of networks are arranged in a hierarchy that mirrors the parse of
the command.
This encodes the command into the structure of the model.
A search tree is created through a modified RRT~\citep{lavalle1998rapidly} which
explores different configurations of the robot and their effect on the
environment.
The search procedure is augmented by the hierarchical network which can
influence the nodes being expanded and the direction of expansion.
As the search tree splits into multiple branches, the hierarchical recurrent
network similarly splits following the tree.
This encodes the reasoning and state of the robot if it were to follow that
specific set of actions.
At each time point, the network predicts the likelihood that the action satisfies
the command.
In the end, much as with a classical sampling-based planner, a search tree is
created that explores different options, and a path in that tree is selected
to be executed.

Robustness to new environments is achieved by trading off the planner against
the hierarchical network.
The influence of the network is proportional to its confidence.
When new obstacles, map features, or other difficulties are encountered (for example,
not immediately seeing a goal), the algorithm can temporarily devolve into a
traditional RRT.
This is a desirable feature because algorithms like RRT make optimal decisions
when other guidance is not available.
Unlike planners, uncertain or untrained networks generally make pathologically
bad decisions in such settings.
This issue is often alleviated with techniques such as $\epsilon$-greedy
learning~\citep{watkins1989learning} which provide arbitrary random moves rather
than the near-optimal exploration that sampling-based planners engage in.

\begin{figure}
  \centering
  \vspace{0.5ex}
  \begin{small}
    \begin{tikzpicture}
      \node[draw,solid,black] (frame) {\includegraphics[width=0.06\textwidth]{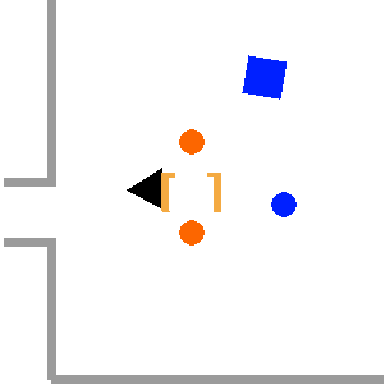}};
      \node[draw,solid,black,left=0.5 of frame] (frame1) {\includegraphics[width=0.06\textwidth]{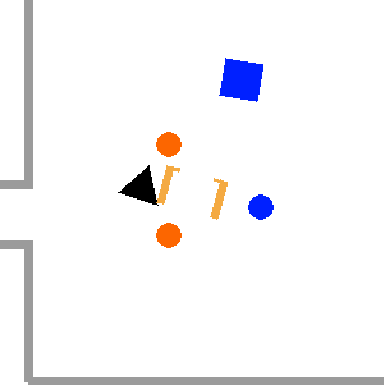}};
      \node[draw,solid,black,left=2.3 of frame] (frame2) {\includegraphics[width=0.06\textwidth]{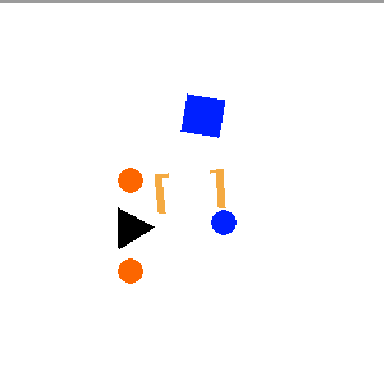}};
      \node[draw,solid,black,right=0.5 of frame] (frame4) {\includegraphics[width=0.06\textwidth]{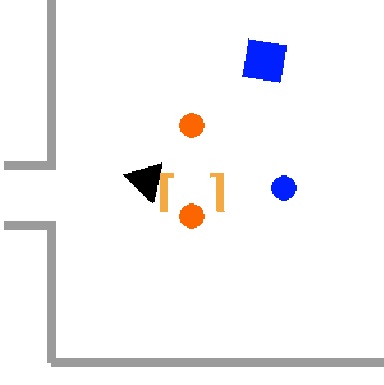}};
      \node[draw,solid,black,right=2.3 of frame] (frame5) {\includegraphics[width=0.06\textwidth]{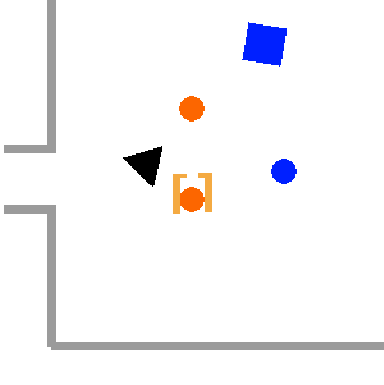}};
      \node[left=-0.1 of frame] () {$\cdots$};
      \node[left=1.7 of frame] () {$\cdots$};
      \node[right=-0.05 of frame] () {$\cdots$};
      \node[right=1.75 of frame] () {$\cdots$};
      \node[above=0.3 of frame] (encoder) {Visual feature CNN};
      \node[above=0.6 of encoder, xshift=10ex,Green] (ring) {Ball};
      \node[above=0.6 of encoder, xshift=-10ex,Green] (triangle) {Triangle};
      \node[above=0.6 of ring,Green] (orange) {Orange};
      \node[above=0.6 of triangle,Green] (black) {Black};
      \node[above=4.0 of encoder] (below) {};
      \node[below=0 of below, xshift=6ex] (RNN) {RNN};
      \node[below=0.2 of RNN] (x) {$\times$};
      \node[below=1.3 of below] (attention) {Attention prediction};
      \node[left=1 of below,Grey,Green] (belowText) {Below};
      \node[above=0.4 of below,Green] (pickup) {Pick up};
      \node[above=0.4 of pickup] (proposal) {Proposal layer};
      \node[above=0.2 of proposal,xshift=-10ex,Grey] (direction) {Direction};
      \node[above=0.2 of proposal,xshift=10ex,Grey] (success) {Stop probability};
      \begin{scope}[on background layer]
        \node (corpus) [rounded corners, dashed, draw=black, fit= (below)(belowText)(attention)(RNN), fill=gray, opacity=0.1,
        inner sep=0.1ex] {};
      \end{scope}
      \draw[-{Latex[length=1.2ex,width=1.2ex]},thin,black] (frame) -- (encoder);
      \draw[-{Latex[length=1.2ex,width=1.2ex]},thin,Blue] (encoder) -- (ring);
      \draw[-{Latex[length=1.2ex,width=1.2ex]},thin,Blue] (encoder) -- (triangle);
      \draw[-{Latex[length=1.2ex,width=1.2ex]},thin,Blue] (encoder) -- (black);
      \draw[-{Latex[length=1.2ex,width=1.2ex]},thin,Blue] (encoder) -- (orange);
      \draw[-{Latex[length=1.2ex,width=1.2ex]},thin,Blue] (encoder) -- (attention);
      \draw[-{Latex[length=1.2ex,width=1.2ex]},thin,Blue] ($(encoder.north)+(0.1,0)$) -- ($(pickup.south)+(0.1,0)$);
      \draw[-{Latex[length=1.2ex,width=1.2ex]},thin,Blue] ($(encoder.north)+(0.1,2.55)$) to[out=90,in=180] ($(x.west)+(0.1,-0.05)$);
      \draw[-{Latex[length=1.2ex,width=1.2ex]},thin,Red] (triangle) -- (black);
      \draw[-{Latex[length=1.2ex,width=1.2ex]},thin,Red] (ring) -- (orange);
      \draw[-{Latex[length=1.2ex,width=1.2ex]},thin,Red] (black) -- (attention);
      \draw[-{Latex[length=1.2ex,width=1.2ex]},thin,Red] (orange) -- (attention);
      \draw[-{Latex[length=1.2ex,width=1.2ex]},thin,Red] (attention) -- (pickup);
      \draw[-{Latex[length=1.2ex,width=1.2ex]},thin,Red] (attention.north) to[out=90,in=180] ($(x.west)+(0.1,0.05)$);
      \draw[-{Latex[length=1.2ex,width=1.2ex]},thin,Red] (pickup) -- (proposal);
      \draw[-{Latex[length=1.2ex,width=1.2ex]},thin,Black] (proposal) -- (direction);
      \draw[-{Latex[length=1.2ex,width=1.2ex]},thin,Black] (proposal) -- (success);
      \draw[-{Latex[length=1.2ex,width=1.2ex]},thin,Black] ($(pickup.north)+(0.2,0)$) -- ($(proposal.south)+(0.2,0)$);
      \draw[-{Latex[length=1.2ex,width=1.2ex]},thin,black] ($(x.north)+(0,-0.1)$) -- (RNN);
      \draw[thin,orange,densely dashed,-{Latex[length=1.2ex,width=1.2ex]}] ($(pickup.west)+(-2.2,0.05)$) -- ($(pickup.west)+(0,0.05)$);
      \draw[thin,orange,densely dashed,-{Latex[length=1.2ex,width=1.2ex]}] ($(pickup.east)+(0,0.05)$) -- ($(pickup.east)+(2.3,0.05)$);
      \draw[thin,orange,densely dashed,-{Latex[length=1.2ex,width=1.2ex]}] ($(RNN.west)-(3.3,0)$) -- (RNN.west);
      \draw[thin,orange,densely dashed,-{Latex[length=1.2ex,width=1.2ex]}] (RNN.east) -- ($(RNN.east)+(1.6,0)$);
      \draw[thin,orange,densely dashed,-{Latex[length=1.2ex,width=1.2ex]}] ($(RNN.west)-(3.3,0)$) to[out=0,in=180] (attention.west);
      \draw[thin,orange,densely dashed,-{Latex[length=1.2ex,width=1.2ex]}] ($(black.west)+(-1,0.05)$) -- ($(black.west)+(0,0.05)$);
      \draw[thin,orange,densely dashed,-{Latex[length=1.2ex,width=1.2ex]}] ($(black.east)+(0,0.05)$) to[out=9,in=171] ($(black.east)+(3.9,0.05)$);
      \draw[thin,orange,densely dashed,-{Latex[length=1.2ex,width=1.2ex]}] ($(orange.west)+(-3.75,-0.05)$) to[out=-9,in=-171] ($(orange.west)+(0,-0.05)$);
      \draw[thin,orange,densely dashed,-{Latex[length=1.2ex,width=1.2ex]}] ($(orange.east)+(0,-0.05)$) -- ($(orange.east)+(0.96,-0.05)$);
      \draw[thin,orange,densely dashed,-{Latex[length=1.2ex,width=1.2ex]}] ($(triangle.west)+(-0.82,0.05)$) -- ($(triangle.west)+(0,0.05)$);
      \draw[thin,orange,densely dashed,-{Latex[length=1.2ex,width=1.2ex]}] ($(triangle.east)+(0,0.05)$) to[out=9,in=171] ($(triangle.east)+(3.7,0.05)$);
      \draw[thin,orange,densely dashed,-{Latex[length=1.2ex,width=1.2ex]}] ($(ring.west)+(-3.9,-0.05)$) to[out=-9,in=-171] ($(ring.west)+(0,-0.05)$);
      \draw[thin,orange,densely dashed,-{Latex[length=1.2ex,width=1.2ex]}] ($(ring.east)+(0,-0.05)$) -- ($(ring.east)+(1.08,-0.05)$);
    \end{tikzpicture}
  \end{small}
  \caption{The structure of the model interpreting and following the command
    \emph{Pick up the orange ball from below black triangle}. As the search tree
    described in \cref{fig:model-overview} is constructed, this model interprets
    the state of each tree node being expanded. It predicts the direction to
    expand the node in and whether the node completes the plan being
    followed. Each {\green word is a module} in the network, each module
    contains two neural networks (shown in black --- the module for \emph{below}
    is expanded). Each word updates its associated {\orange hidden state}
    updated at each time step using an RNN. The structure of the network is
    derived automatically from a parse produced by the NLTK coreNLP
    parser~\citep{bird2006nltk}. {\blue Visual features} are extracted and
    provided to each word model. {\red Attention maps} are predicted by each
    word by a combination of visual features, the attention maps of any words
    directly below in the hierarchy, and the state of that word. The attention
    maps indicate which objects should be manipulated and how they should be
    manipulated. The attention map of the final word and the output of its RNN
    are used to predict the direction of movement and the success
    probability. Using attention maps as the mechanism to forward information in
    the network provides a level of interpretability.}
  \label{fig:model}
  \label{fig:rnn-model}
\end{figure}

We ensure that the model provides a level of interpretability in two ways.
First, the structure of the sentence is encoded explicitly into the structure of
the recurrent network; see~\cref{fig:rnn-model}.
Inspecting the network reveals which subnetworks are connected together and the
topology of the connections mirrors that of natural language.
Second, the internal reasoning of the model is highly constrained to operate
through attention maps.
Rather than allowing each component the freedom to pass along any information up
the hierarchy in order to make a decision, we constrain all components to
communicating via a grayscale map that is multiplied by the current observation
of the environment.
Inspecting these attention maps reveals information about which areas each
network is focused on and can provide a means to understand and explain
failures.
In addition, this constrained representation is easy to learn and does not
require a large number of examples.
We find that adding these interpretable computations also increases performance
relative to more opaque representations, likely because words which have never
co-occurred at training time have an easier time understanding the output of
other word models when the representations are interpretable.

This work makes four contributions.
\begin{enumerate}
\item We demonstrate how a robotic planner, in addition to reasoning about
  physical affordances and obstacles, can be extended to reason about a linguistic
  command in a continuous environment.
\item We show that a hierarchical recurrent model structured according to the
  parse of a sentence can learn the meanings of sentences efficiently thereby
  guiding a robot's motion and manipulation.
\item We demonstrate that such a model generalizes to new settings, to more
  challenging maps that include obstacles not seen in the training set, and to
  longer commands.
\item By constraining the model to reason visually through attention maps rather
  than arbitrary vectors, we produce an end-to-end model with more interpretable
  intermediate reasoning steps without needing intermediate symbolic
  representations.
\end{enumerate}

\section{RELATED WORK}
\label{sec:related-work}

Much prior work has explored how symbolic representations in high-level planning
languages, such as PDDL~\citep{fox2003pddl2}, can ground linguistic commands.
In addition some approaches combine together task planning and motion planning
using symbolic representations for the
task~\citep{kaelbling2013integrated,garrett2018ffrob,dantam2018task,eppe2019semantics}.
Such approaches can plan efficiently in large continuous spaces but require a
symbolic representation of a task, must be manually created, cannot be trained
to acquire new concepts, and do not handle ambiguity well.
Our approach similarly combines task and motion planning, but does so without
symbolic representations albeit with simpler tasks than state-of-the-art models
in such domains can handle. 
Prior work has built models that can robustly follow linguistic commands on top
of these architectures~\citep{tellex2011understanding,paul2017temporal}.
Because of the underlying symbolic nature of the final representation, this prior work
cannot acquire concepts that are not easily expressed in the target planning
language and cannot learn new primitives in that language.
Paxton~\etal~\citep{paxton2019prospection} demonstrate a model which breaks down
tasks for such planners automatically --- it learns to map sentences to a
sequence of subgoals.
Bisk~\etal~\citep{bisk2016natural} demonstrate how to break down manipulation tasks and how
to ground them to perception from natural language input, much like the tasks we
perform here, but do not execute such commands.

Fu~\etal~\citep{fu2019language} and Shah~\etal~\citep{shah2018follownet}
demonstrate an approach which can map sentences to robotic actions (navigation
\& pick and place) via multi-task reinforcement learning in a discrete state and
action space.
This is similar in spirit to our approach but we do so in continuous action and
state spaces which require many precise steps in the configuration space to
execute what would otherwise be a single output token such as ``pick up'' for
discrete problems.

Blukis~\etal~\citep{Blukis:18visit-predict} demonstrate how a drone can be
controlled by predicting the goal configuration of the robot.
Their model operates in a continuous space but does not contain object
interactions, manipulations, or obstacles.
Predicting a single final goal for such complex multistep actions is infeasible,
as the goal must contain not just the position of the robot but the position of
the other objects.

\section{DEEP PLANNERS WITH LANGUAGE}

We describe how an extension of RRT adds neural networks that control the search
process, \cref{sec:deep-rrt}, then describe the structure of the model used and
how it encodes complex sentences, \cref{sec:lang-rrt}, and finally how such
networks can be efficiently trained, \cref{sec:train-lang-rrt}.

\subsection{Planning with Deep RRT}
\label{sec:deep-rrt}

Robotic planners are efficient at searching the configuration spaces of robots.
Here we describe how to augment them with networks that efficiently learn
language, how to guide the planning process, and how to recognize when a plan
described by a sentence has been completed.
This is related to the approach of Kuo \etal~\citep{kuo2018deep}, DeRRT, which
introduced deep sequential models for sampling-based planning.
It took an RRT-based planner and described how to guide its behavior with a
neural network.

The planner maintains a search tree and a corresponding recurrent neural network
with the goal of reaching a fixed destination in the configuration space.
This is related to the model shown in \cref{fig:model-overview} --- in this work,
we use a collection of networks and have the networks determine the final
configuration based on the command rather than explicitly providing it.
The goal of a traditional planner is to construct a search tree that explores
the space and connects the start state to an end state.
It will then choose the best path between the two from the tree for the robot to
follow.

At each step, the planner chooses a node to extend by the standard RRT mechanism:
sampling a point in space and finding the nearest point in the tree to that
sample.
It then proposes a new tree node between the selected tree node and the sampled
point.
The neural network takes as input the state at the current node, any visual
observations at the current node, and the proposed extension to the tree.
Observations are processed with a co-trained CNN that is shared among all words.
The network makes its own prediction about how to extend the tree at the current
node along with a confidence value.
A simple mixture model selects between the planner and the network proposed
directions.
Once the tree is constructed, in this case after a fixed number of planning
steps, the node which is considered most likely to be an end state of the
described command is chosen and the path between the start state and that node
is generated.

We modify this algorithm to influence the choice of which node to expand, not
just the direction to expand a preselected node in.
These modifications can be applied to the original version of the algorithm
presented in that earlier work.
The neural network that guides the planner is trained to maximize the likelihood
of the data provided.
This results in a probability assigned to each node and each path.
Every search tree node is annotated with its likelihood conditioned on its
parent as well was with the likelihood of being chosen for expansion by the
vanilla RRT.
The latter probability is computed by generating many free-space samples and
computing the distribution over which nodes are extended --- this is very
computationally efficient.
To then sample which node to extend, we multiply and normalize these
probabilities, sampling from the distribution of nodes which would be chosen by
both the network and RRT.
This focuses search in areas where plans are likely to succeed while not
allowing the neural network to get stuck in one region --- as a region is more
saturated with samples the likelihood that RRT would continue to extend the tree
in it becomes very low.

\subsection{Language and Deep RRT}
\label{sec:lang-rrt}

The model described thus far uses a single recurrent network in order to guide
the planner.
Technically, this is serviceable, as the network can in principle learn to
perform this task.
Practically, generalizing to new sentences and complex sentential structures is
beyond the abilities of that simple model.

Using a collection of networks, rather than a single network can help
generalization to new sentences.
Just as one network can guide a planner, multiple networks can also guide it.
Each network can make a prediction.
A direction can be sampled from the posterior distribution over all of the
predictions by all of the networks.

We build this collection of networks out of a lexicon of component networks.
Given a training set of commands and paths, one component network is trained per
word in the command that the robot is following.
Given a test sentence, the words in the sentence determine the set of component
networks which guide the planner.
We call this the \emph{bag of words}, BoW, model because there is no explicit
relationship or information flow between the networks.
Due to the lack of relationships between words, this model has fundamental
difficulties representing the difference between \emph{Grab the black toy from
  the box} and \emph{Grab the toy from the black box}.

To address this limitation, we introduce a hierarchical network; see~\cref{fig:rnn-model}.
Given a sentence and the parse of the sentence derived from the NLTK coreNLP
parser~\citep{bird2006nltk}, we select the same set of component networks that
correspond to the words in the sentence.
Networks are arranged in a tree --- naturally, such trees are rooted by verbs in
most linguistic representations.
The state at the current node informs the representation of each component
network.
Each component updates its own hidden state and forwards information to all of the
components that are linked to it.
The leaves only receive as input observation at the current state and their own
hidden state.
The root of the tree produces an output used by a linear proposal layer to
predict the direction of movement and the likelihood that the current node has
reached a goal.
This approach has the ability to represent the earlier distinction about which
noun the word \emph{black} is attached to because different attachments result in
different parse trees and thus different instantiations of the model.
One must pay particular attention to argument structure here -- verbs which take
multiple arguments such as \emph{give} must always take them in the same order
(the object of the given and the destination of the \emph{give} should always
fill the slots of \emph{give}).

We restrict nodes to communicating via attention maps rather than arbitrary
vectors.
This helps generalization as words which never co-occurred in the training set
can be seen in the test set of the experiments we report.
By ensuring that the representation shared between component networks is
universal, such as the attention maps, component networks are encouraged to be
more compatible with one another.
This is enforced by the structure of each component network, \ie each network
corresponding to a word.
Each word takes as input a set of attention maps, weighs the input image with
each attention map independently, and combines this with the hidden state of that word.
A new attention map is predicted and passed to subsequent words.
Using this predicted attention map, an RNN takes as input the observed image
weighted by the attention map and updates the hidden state of the word.
In addition to encouraging generalization, attention maps can be interpreted by
humans, and help speed up learning by being relatively low dimensional.

\subsection{Training Compositional Deep RRT}
\label{sec:train-lang-rrt}

There are three parts of the model that must be trained: the shared CNN that
embeds visual observations, a lexicon of component networks, and the proposal layer.
The lexicon of component networks maps words to networks that represent the
meanings of those words.
We might in principle annotate when a word is relevant to a plan and train each
word independently, but we find that joint training allows easier and
already-known words to supervise new words.
This is because the hierarchical nature of the model allows information flow
between words, giving words which have high confidence an opportunity to help
guide the representation of words which are not yet well trained.
Instead, we train the model with little supervision: pairs of sentences and
paths.
The model is not informed about which parts of the sentence correspond to which
parts of the path, when words are relevant, or how words relate to visual
observations.

The overall model is trained in two phases. First, all weights are
trained including the shared CNN that embeds visual observations, the lexicon of
component networks, and the direction in which to extend the search tree.
Next, these three sets of weights are fixed while the proposal layer is
fine-tuned to predict the likelihood of a state being a goal state of a plan.
This fine-tuning step significantly increase performance without requiring
more training data: the proposal layer gains experience with how to
interpret the output of the network without the network also changing its
behavior.

The model presented here and summarized in~\cref{fig:model-overview}
and~\cref{fig:rnn-model} is trained with little data, only sentences paired with
demonstrations.
It operates efficiently in continuous environments.
Its structure is intelligible and derived from linguistic principles while its
reasoning is made overt by the explicit use of attention maps.
Next, we describe how we evaluate this model.

\section{EXPERIMENTS}

To evaluate the model, we describe the task and training set generation
procedure, \cref{sec:ex-dataset}, then describe baseline models,
\cref{sec:ex-models}, test the ability of the model to carry out novel commands,
\cref{sec:ex-commands}, the ability to generalize to novel features in the
environment, \cref{sec:ex-obstacles}, generalize to multiple sentences,
\cref{sec:multiple-sentences}, and the ability to handle real-world commands
generated by users, \cref{sec:user-study}.

\subsection{Dataset}
\label{sec:ex-dataset}

\begin{figure}
  \vspace{2ex}
  \begin{small}
    \begin{tabular}{c@{\hspace{1ex}}c@{\hspace{1ex}}c@{\hspace{1ex}}c}
      \includegraphics[width=0.23\linewidth]{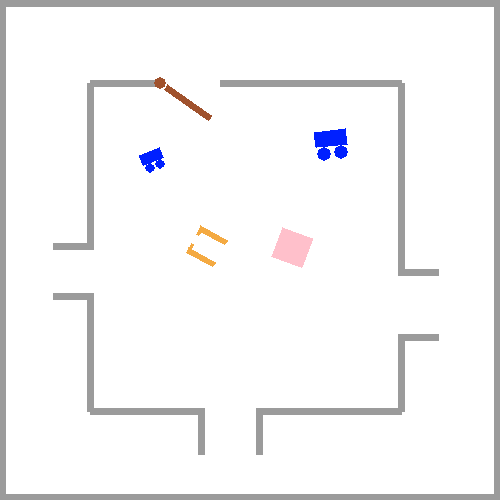}&
      \includegraphics[width=0.23\linewidth]{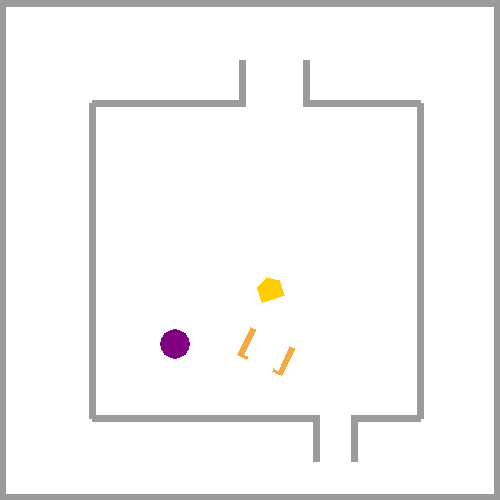}&
      \includegraphics[width=0.23\linewidth]{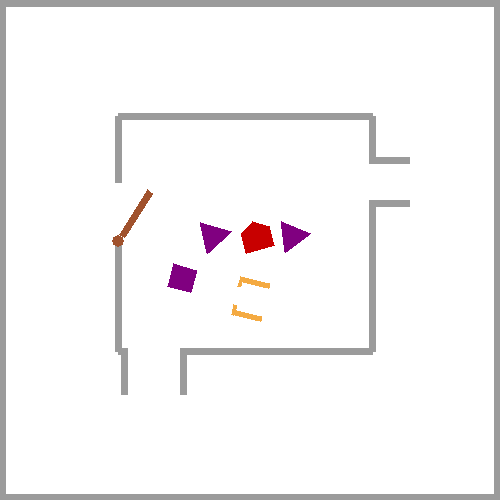}&
      \includegraphics[width=0.23\linewidth]{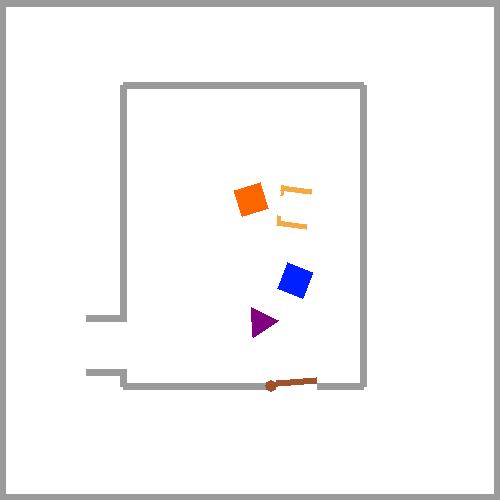}\\
      \multicolumn{4}{c}{(a) Representative examples of the training set}\\
      \includegraphics[width=0.23\linewidth]{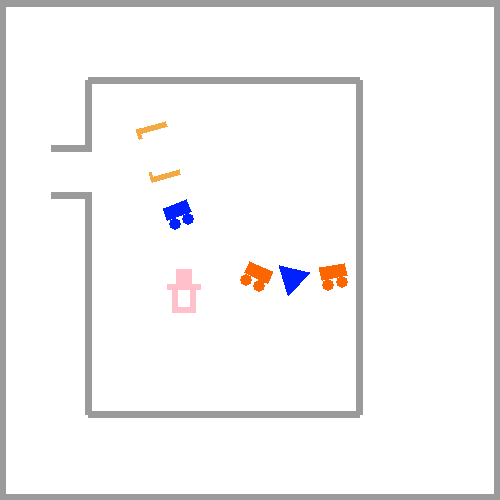}&
      \includegraphics[width=0.23\linewidth]{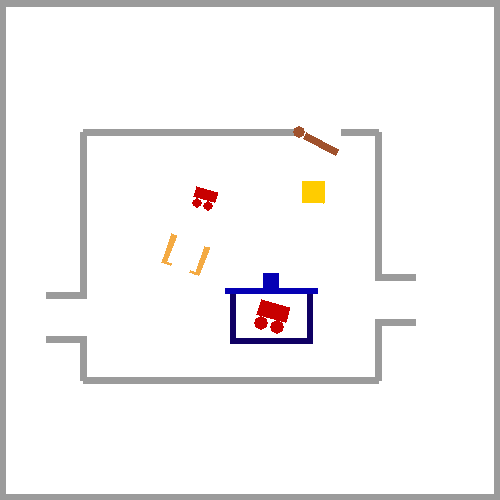}&
      \includegraphics[width=0.23\linewidth]{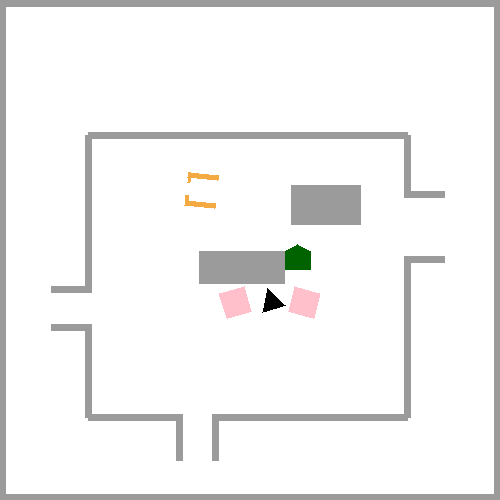}&
      \includegraphics[width=0.23\linewidth]{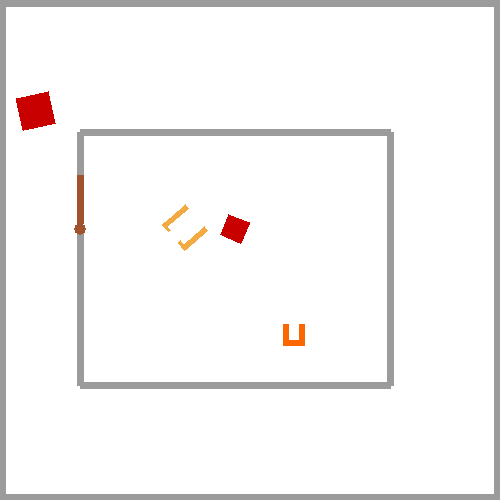}\\
      \multicolumn{4}{c}{(b) Representative examples of the test set}\\
    \end{tabular}
  \end{small}
  \caption{Examples of the (a) training set and of the (b) test set. The
    {\orange robot} is shown in orange as a pair of L-shaped grippers. Objects
    are randomly positioned, with random properties and orientations. The
    training set is considerably simpler, with fewer objects on average, without
    cups that have lids, without the need to traverse doors or channels as all
    objects are inside the room, and without immovable obstacles (grey
    rectangles).}
  \label{fig:map-examples}
\end{figure}

A generative model creates new training and test maps conditioned on a target
command which is sampled from a grammar.
The space of possible maps is large and includes rooms of varying sizes, which
can have between 0 and 4 narrow gaps, possibly contain a door to the outside,
and may contain between two and eight objects with multiple properties (shape,
color, and size); see~\cref{fig:map-examples}.
The grammar which generates commands contains seven verbs (push, grab, approach,
touch, open, leave, carry), seven nouns (block, cup, ball, triangle,
quadrilateral, house, cart), eight colors (red, green, blue, pink, yellow,
black, purple, orange), two sizes (big, small), nine spatial relations (left of,
right of, top of, bottom of, on the left of, on the right of, near, above,
below), and two prepositions (towards, away from).
Each of these linguistic constituents becomes a component neural network in a
lexicon of networks.
Sentences are parsed with the NLTK coreNLP parser~\citep{bird2006nltk} and
unknown words in the sentences are mapped to nearby words using their distance
in WordNet~\citep{miller1995wordnet,pedersen2004wordnet}.

Given all of the possible objects, distractors, room sizes, doors, gateways,
object locations, color, rotation, and size, a random map is generated.
We verify that the target plan is in principle feasible on this map.
The same map never appears in both the training and the test sets.
This provides an immense space from which to generate maps and to test model
generalization capabilities.

\subsection{Models}
\label{sec:ex-models}

As described in the related work, \cref{sec:related-work}, no existing model is
able to take as input linguistic commands and plan in the environments used
here.
To evaluate our model, we develop several baselines by augmenting earlier work
to perform this challenging task.

The weakest baseline, \emph{RNN-Only}, is our model without the planner but
including the hierarchical neural network.
A more powerful baseline, \emph{BoW}, is created by augmenting the work
in Kuo \etal \citep{kuo2018deep}.
The network described there is given the added task of predicting when a
configuration of the robot is a terminal.
A collection of neural networks represent the meaning of a sentence, but they do not
interact with one another; these form the \emph{bag of words}.
This model is novel and related to our own, but considerably weaker as there is
no relationship between the words and no explicit encoding of the structure of
the sentence.
The neural network used in that is also modified to predict both a direction to
move in and the probability of ending the action --- similarly to the model
presented here but using a single neural network.

Finally, we compare against a model, \emph{RRT+Oracle}, which represents the
performance that can be expected if the hierarchical network is operating
well.
This model uses the same underlying planner but the goal regions are manually
specified through the use of an oracle.
For any position in the configuration space of the robot and the configuration
space of all of the objects, the oracle determines if the behavior of the robot
has satisfied some natural language utterance.
Equaling this strong related model in performance demonstrates that the network
is acquiring the meanings of words.

\subsection{Understanding Natural Language Commands}
\label{sec:ex-commands}

\begin{figure}
  \centering
  \vspace{0.7ex}
  \scalebox{1.1}{%
  \begin{scriptsize}
    \begin{tabular}{ccc}
      Planner       & \multicolumn{1}{c}{Two concepts} & \multicolumn{1}{c}{Five or six concepts}\\
      RNN-Only      & 0.25                             & 0.24    \\
      BoW           & 0.61                             & 0.36    \\
      \textbf{Ours} & \textbf{0.72}                    & \textbf{0.50}  \\
      RRT+Oracle     & 0.64                             & 0.49    \\
    \end{tabular}
  \end{scriptsize}}
  \caption{Success rate of executing natural language commands with two
    concepts, the same number as the models saw at training time, and five or
    six concepts, more complex sentences than were seen at training time. All
    models sampled 500 nodes in the configuration space of the robot. Our model
    generalizes well and faithfully encodes the meaning of commands. While the
    BoW model is also novel it lacks the internal structure to represent many
    sentences and significantly underperforms our hierarchical model.}
  \label{fig:learning}
\end{figure}

First, we test if our model can acquire the meanings of words and use this to
represent never-before-seen sentences; see~\cref{fig:learning}.
Note that very little is annotated here: only pairs of demonstrations and
sentences related to those demonstrations exist.
Also note that for all experiments, test training and test maps and utterances
were disjoint.

We generated a training set of 6099 utterances containing at most four concepts,
with each utterance being demonstrated on a new map.
The test set consisted of 657 utterances paired with maps that do not appear in
the training set and are on the whole considerably more complex; see~\cref{fig:map-examples}.
The model presented here had by far the highest success rate (72\%) and
generalized best to more complex sentences.
At training time, we never presented sentences that had more than four concepts,
while at test time, we included considerably more complex sentences.
Our model generalized to these longer sentences despite not having seen anything
like them at training time.

Since our model affects the search direction of RRT, \ie the growth of the
search tree, it even outperformed the \emph{RRT+Oracle} model.
The \emph{RRT+Oracle} model has a perfect understanding of the sentence in terms
of determining which nodes satisfy the command, but lacks the ability to use the
sentence to guide its actions.
This demonstrates that the model presented here faithfully encodes commands and
executes them well in complex environments, on new maps, even when those
commands are much more complex than those seen in the training set.

\subsection{Additional Obstacles and Preconditions}
\label{sec:ex-obstacles}

\begin{figure}
  \centering
  \vspace{1ex}
  \scalebox{1.1}{%
  \begin{scriptsize}
    \begin{tabular}{cccc}
      Planner & \multicolumn{1}{c}{Obstacles} & \multicolumn{1}{c}{Cup \& Lid} & \multicolumn{1}{c}{Door}\\
      RNN-Only      & 0.12          & 0.08          & 0             \\
      BoW           & 0.32          & 0.08          & \textbf{0.35} \\
      \textbf{Ours} & \textbf{0.52} & \textbf{0.16} & 0.30          \\
      RRT+Oracle    & \textbf{0.52} & 0.08          & \textbf{0.35} \\
    \end{tabular}
  \end{scriptsize}}
  \caption{The success rate of different baselines and models when generalizing
    to environments that have properties which have never or rarely been
    experienced at training time. Note that, the \emph{RNN-Only} model which
    does not include a planner, fails to generalize. Models which do include a
    planner generalize much better to new problems.}
  \label{fig:ex-obstacles}
\end{figure}

Robots must continually deal with new difficulties.
To evaluate the capacity of models to adapt to new problems, we further modify
the test set to include other features not present at training time;
see~\cref{fig:ex-obstacles}.
In particular, we add up to four random fixed obstacles and require that the
robot traverse a push-button-controlled door.
In addition, the frequency of objects inside cups with lids is significantly
increased.

The model which does not include a planner, the RNN-Only model, has great
difficulty generalizing to new scenarios.
All the other models generalized far better, with ours performing roughly on par
with the oracle.
These results indicate that planners provide robustness when encountering new
challenges.
This is well known in symbolic planning but has not been exploited as part of an
end-to-end approach before.

\subsection{Multiple Sentences}
\label{sec:multiple-sentences}

\begin{figure}
  \vspace{0.7ex}
  \centering
  \scalebox{1.1}{%
  \begin{scriptsize}
    \begin{tabular}{ccccc}
      Planner       & \multicolumn{4}{c}{Number of sentences}\\ \cmidrule(lr){2-5}
                    & 1             & 2             & 3             \\[0.5ex]
      RNN-Only      & 0.17          & 0             & 0             \\
      BoW           & 0.40          & 0.12          & 0.06          \\
      \textbf{Ours} & \textbf{0.58} & \textbf{0.33} & 0.10          \\
      RRT+Oracle    & 0.52          & 0.25          & \textbf{0.12} \\
    \end{tabular}
  \end{scriptsize}}
  \caption{All models are trained on a single utterance and are then required to
    follow a sequence of commands. Every model is allowed to sample 600 nodes in
    the configuration space of the robot. As more commands are added, carrying
    out a task becomes increasingly difficult and the RNN-Only model is quickly
    overwhelmed. The BoW model performs at roughly half of the performance of
    ours. Our model has performance comparable to that of the oracle. Sampling
    more nodes would increase the success rate of all models.}
  \label{fig:multiple-sentences}
\end{figure}

Robots are unlikely to be required to carry out just one command at a time.
Most plans will include a sequence of actions that depend on one another.
We choose to evaluate an extreme version of this task where all models are only
trained on a single sentence and then must generalize to sequences of between
two and three commands; see~\cref{fig:multiple-sentences}.
Despite this significant limitation at training time, we find that our approach
outperforms the baselines significantly.
The RNN-Only model is unable to generalize.
The BoW model has roughly half the performance of our model.
Our model has similar properties to that of the oracle, which
has the correct encoding of the sentence, thus showing that our model represents
sequences of sentences despite not being trained on any sequences of commands.

\subsection{User Study}
\label{sec:user-study}

We generated 500 map and command pairs and had the robot execute those commands.
The executions of these commands, but not the commands themselves, were shown to
four users recruited for this experiment.
Users were asked to produce the instructions they would provide to the robot in
order to elicit the behavior they observed.
Out of 500 descriptions, 128 were impossible for the robot to follow due to user
error, \eg by mentioning objects that are physically not there, or could not be
reasonably parsed.
The 372 remaining descriptions had an average length of 9.04 words per sentence
standard deviation of 2.49.
The baseline RNN-Only model achieved 17\% success rate, the BoW model succeeded
40\% of the time, while our model succeeded 49\% of the time.
The RRT+Oracle model had roughly the same performance as ours succeeding 51\% of
the time.
This demonstrates that our approach scales to real-world user input.

\section{DISCUSSION \& CONCLUSION}

We have demonstrated that a hierarchical recurrent network can work in
conjunction with a sampling-based planner to create a model that encodes the
meaning of commands.
It learns to execute novel commands in challenging new environments that contain
features not seen in the training set.
We demonstrated that our approach scales to real-world sentences produced by
users.

\begin{figure}
  \centering
  \vspace{2ex}
  \begin{small}
    \begin{tabular}{cccc}
      \includegraphics[width=0.2\linewidth,frame]{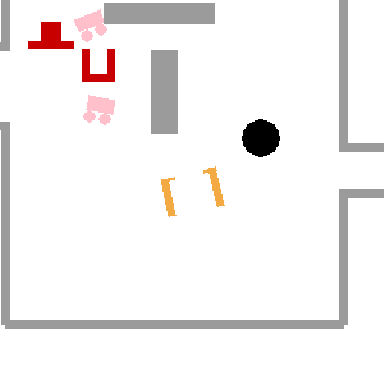}
      &\includegraphics[width=0.2\linewidth,frame]{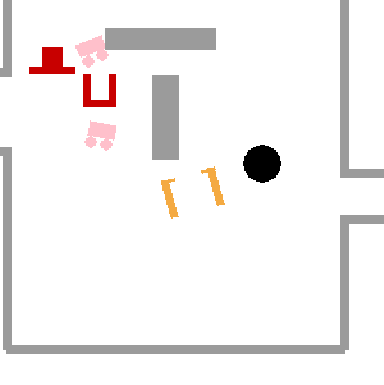}
      &\includegraphics[width=0.2\linewidth,frame]{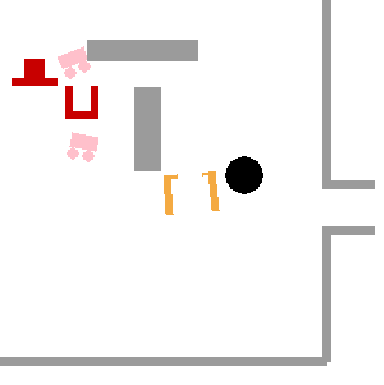}
      &\includegraphics[width=0.2\linewidth,frame]{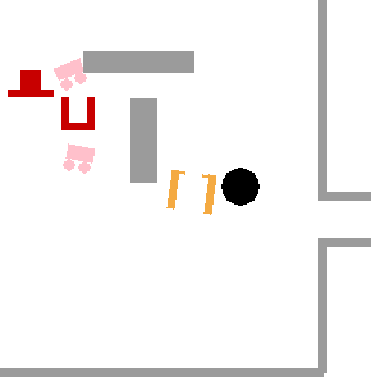}\\
      \includegraphics[width=0.2\linewidth]{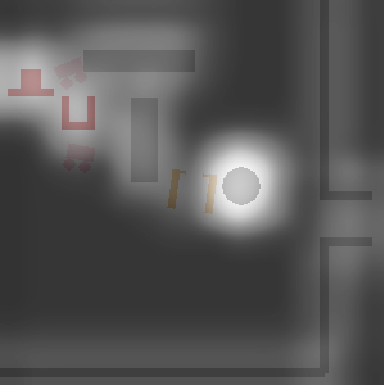}
      &\includegraphics[width=0.2\linewidth]{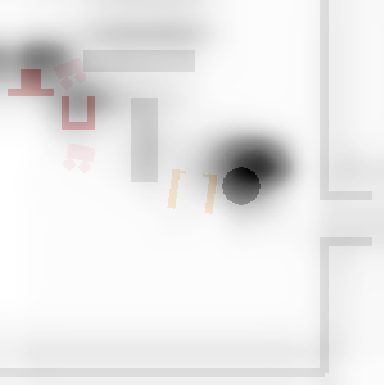}
      &\includegraphics[width=0.2\linewidth]{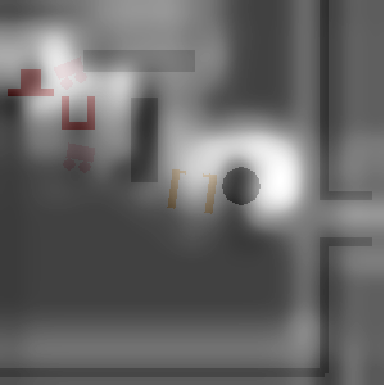}
      &\includegraphics[width=0.2\linewidth]{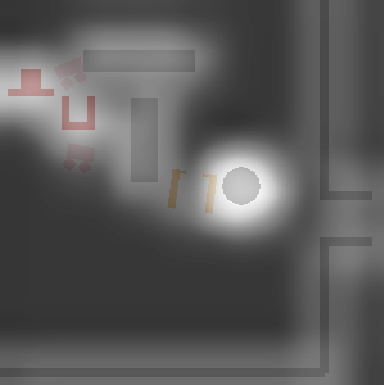}\\
      \emph{cup} & \emph{cart} & \emph{above} & \emph{approach}\\
      \multicolumn{4}{c}{(a) {\red Failure}: Approach the cart above the cup}\\[2ex]
      \includegraphics[width=0.2\linewidth,frame]{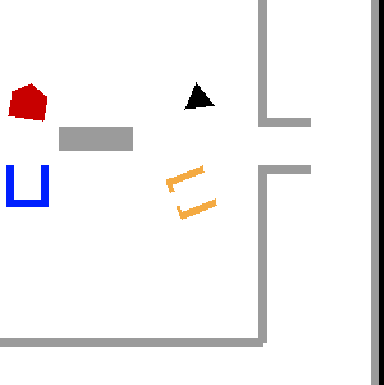}
      &\includegraphics[width=0.2\linewidth,frame]{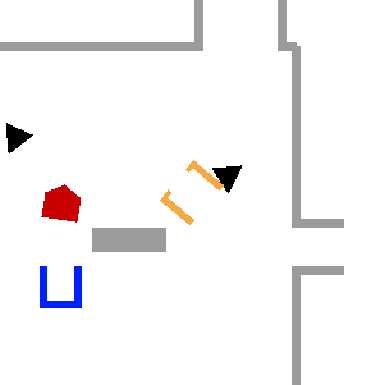}
      &\includegraphics[width=0.2\linewidth,frame]{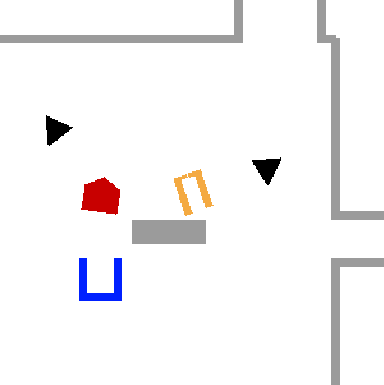}
      &\includegraphics[width=0.2\linewidth,frame]{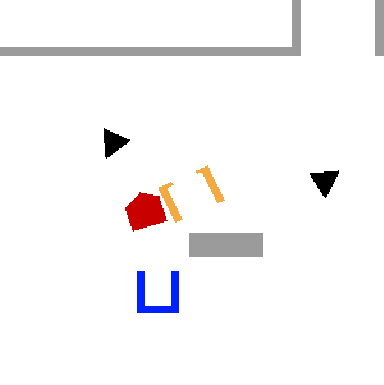}\\
      \includegraphics[width=0.2\linewidth]{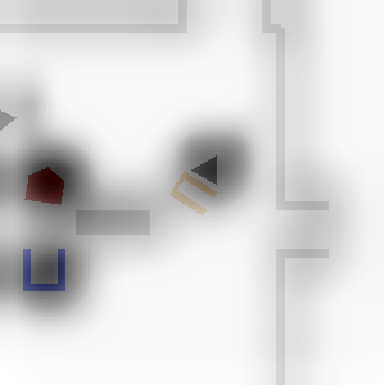}
      &\includegraphics[width=0.2\linewidth]{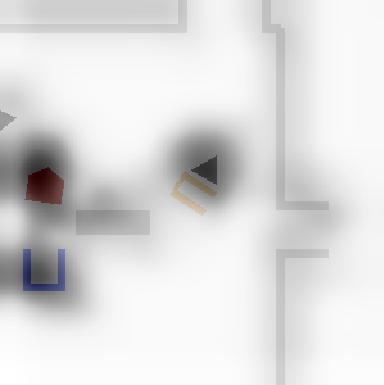}
      &\includegraphics[width=0.2\linewidth]{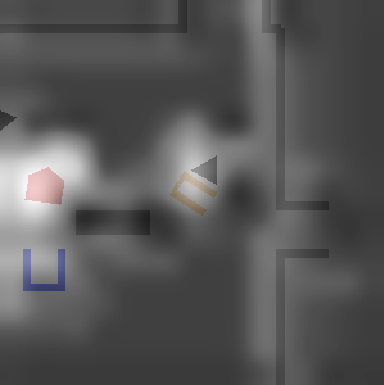}
      &\includegraphics[width=0.2\linewidth]{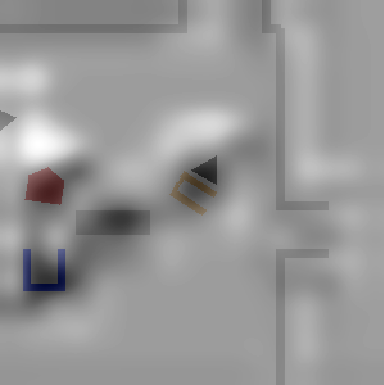}\\
      \emph{triangle} & \emph{house} & \emph{toward} & \emph{carry}\\
      \multicolumn{4}{c}{(b) {\red Failure}: Carry the triangle toward the house}\\
    \end{tabular}
  \end{small}
  \caption{Examples of snapshots from the execution of the model (top; these are
    from the point of view of the robot and represent the local information
    available to it rather than the entire map) along with the attention maps
    (bottom) produced by each component network in the model at a critical time
    in the execution of two commands which were not carried out correctly.  Note
    that the polarity of the attention maps is irrelevant --- models can
    communicate by either suppressing or highlighting features and neither
    carries any a priori valence. In (a) the robot fails to pick up the correct
    object and heads to the circle instead. The failure in (a) is explained by
    the poor detection of the cup seemingly confusing it with the circle. In (b)
    the model goes to the triangle but then fails to pick it up before heading
    to the house. The failure is not explained by the attention maps, as they
    correctly highlight the relevant objects --- instead the model seems to have
    positioned itself incorrectly to perform the pick up (which is attempted)
    and it does not recognize the failure of that action. This level of
    interpretability is not perfect for every failure case, but does explain
    many problems pointing the way for how to improve the model and its training
    regime.}
  \label{fig:attention-failure}
\end{figure}

Our model provides a level of interpretability.
The structure of the model overtly mirrors that of the parse of a sentence
making it easy to verify if a sentence has been incorrectly encoded.
Attention maps are used throughout the hierarchical network to allow component
parts to communicate with one another.
These provide another means by which to understand which components caused a
failure; see \cref{fig:attention-failure} for example attention maps for failed
commands and the level of explanation possible along with its limitations.
In many cases, this provides both reassurances that errors will be pinpointed
to the responsible part of the model and confidence in the chosen model.
This level of transparency is unusual for end-to-end models in robotics.

In the future, we plan to extend the language learning capacities of the
approach and explore ways to carry out more complex tasks.
Thus far we have used an existing off-the-shelf parser, but learning to
interpret the sentences while learning to execute commands can be extremely
useful --- for example it can help disambiguate domain-specific language.
In the future we intend to integrate with such
work~\citep{artzi2013weakly,williams2018learning,ross2018grounding} to acquire
the structure of language, not just the meanings of words and sentences.
The tasks carried out here are concrete and clear, rather than the abstract
tasks, \eg \emph{set up this meeting room}, that robots should ultimately strive
for.
We plan to explore ways to break down tasks into components and keep track of
the temporal relations between components in the future.




\bibliographystyle{IEEEtran}
\renewcommand*{\bibfont}{\footnotesize}
\bibliography{reference}

\begin{thebibliography}{10}
\providecommand{\url}[1]{#1}
\csname url@rmstyle\endcsname
\providecommand{\newblock}{\relax}
\providecommand{\bibinfo}[2]{#2}
\providecommand\BIBentrySTDinterwordspacing{\spaceskip=0pt\relax}
\providecommand\BIBentryALTinterwordstretchfactor{4}
\providecommand\BIBentryALTinterwordspacing{\spaceskip=\fontdimen2\font plus
\BIBentryALTinterwordstretchfactor\fontdimen3\font minus
  \fontdimen4\font\relax}
\providecommand\BIBforeignlanguage[2]{{%
\expandafter\ifx\csname l@#1\endcsname\relax
\typeout{** WARNING: IEEEtran.bst: No hyphenation pattern has been}%
\typeout{** loaded for the language `#1'. Using the pattern for}%
\typeout{** the default language instead.}%
\else
\language=\csname l@#1\endcsname
\fi
#2}}

\bibitem{Blukis:18visit-predict}
V.~Blukis, D.~Misra, R.~A. Knepper, and Y.~Artzi, ``Mapping navigation
  instructions to continuous control actions with position visitation
  prediction,'' in \emph{Proceedings of the Conference on Robot Learning},
  2018.

\bibitem{misra2018mapping}
D.~K. Misra, A.~Bennett, V.~Blukis, E.~Niklasson, M.~Shatkhin, and Y.~Artzi,
  ``Mapping instructions to actions in 3d environments with visual goal
  prediction,'' in \emph{Proceedings of the 2018 Conference on Empirical
  Methods in Natural Language Processing}, 2018.

\bibitem{shah2018follownet}
P.~Shah, M.~Fiser, A.~Faust, J.~C. Kew, and D.~Hakkani-Tur, ``{FollowNet}:
  Robot navigation by following natural language directions with deep
  reinforcement learning,'' \emph{arXiv preprint arXiv:1805.06150}, 2018.

\bibitem{Hsu1997:NarrowPassage}
D.~Hsu, J.-C. Latombe, and R.~Motwani, ``Path planning in expansive
  configuration spaces,'' in \emph{ICRA}, 1997.

\bibitem{Karaman2011:rrtstar}
S.~Karaman and E.~Frazzoli, ``Sampling-based algorithms for optimal motion
  planning,'' \emph{{The International Journal of Robotics Research}}, vol.~30,
  no.~7, pp. 846--894, 2011.

\bibitem{chen2019learning}
B.~Chen, B.~Dai, and L.~Song, ``Learning to plan via neural
  exploration-exploitation trees,'' \emph{arXiv preprint arXiv:1903.00070},
  2019.

\bibitem{kuo2018deep}
Y.-L. Kuo, A.~Barbu, and B.~Katz, ``Deep sequential models for sampling-based
  planning,'' in \emph{2018 IEEE/RSJ International Conference on Intelligent
  Robots and Systems (IROS)}.\hskip 1em plus 0.5em minus 0.4em\relax IEEE,
  2018, pp. 6490--6497.

\bibitem{lee2018gated}
L.~Lee, E.~Parisotto, D.~S. Chaplot, E.~Xing, and R.~Salakhutdinov, ``Gated
  path planning networks,'' in \emph{Proceedings of the 35th International
  Conference on Machine Learning (ICML 2018)}, 2018.

\bibitem{tellex2011understanding}
S.~Tellex, T.~Kollar, S.~Dickerson, M.~R. Walter, A.~G. Banerjee, S.~Teller,
  and N.~Roy, ``Understanding natural language commands for robotic navigation
  and mobile manipulation,'' in \emph{Twenty-Fifth AAAI Conference on
  Artificial Intelligence}, 2011.

\bibitem{paul2017temporal}
R.~Paul, A.~Barbu, S.~Felshin, B.~Katz, and N.~Roy, ``Temporal grounding graphs
  for language understanding with accrued visual-linguistic context,'' in
  \emph{Proceedings of the 26th International Joint Conference on Artificial
  Intelligence}.\hskip 1em plus 0.5em minus 0.4em\relax AAAI Press, 2017, pp.
  4506--4514.

\bibitem{lavalle1998rapidly}
S.~M. LaValle, ``Rapidly-exploring random trees: A new tool for path
  planning,'' 1998.

\bibitem{watkins1989learning}
C.~Watkins, ``Learning form delayed rewards,'' \emph{Ph. D. thesis, King's
  College, University of Cambridge}, 1989.

\bibitem{bird2006nltk}
S.~Bird, ``{NLTK}: the natural language toolkit,'' in \emph{Proceedings of the
  COLING/ACL on Interactive presentation sessions}.\hskip 1em plus 0.5em minus
  0.4em\relax Association for Computational Linguistics, 2006, pp. 69--72.

\bibitem{fox2003pddl2}
M.~Fox and D.~Long, ``{PDDL2}. 1: An extension to {PDDL} for expressing
  temporal planning domains,'' \emph{Journal of artificial intelligence
  research}, vol.~20, pp. 61--124, 2003.

\bibitem{kaelbling2013integrated}
L.~P. Kaelbling and T.~Lozano-P{\'e}rez, ``Integrated task and motion planning
  in belief space,'' \emph{The International Journal of Robotics Research},
  vol.~32, no. 9-10, pp. 1194--1227, 2013.

\bibitem{garrett2018ffrob}
C.~R. Garrett, T.~Lozano-Perez, and L.~P. Kaelbling, ``Ffrob: Leveraging
  symbolic planning for efficient task and motion planning,'' \emph{The
  International Journal of Robotics Research}, vol.~37, no.~1, pp. 104--136,
  2018.

\bibitem{dantam2018task}
N.~T. Dantam, S.~Chaudhuri, and L.~E. Kavraki, ``The task-motion kit: An open
  source, general-purpose task and motion-planning framework,'' \emph{IEEE
  Robotics \& Automation Magazine}, vol.~25, no.~3, pp. 61--70, 2018.

\bibitem{eppe2019semantics}
M.~Eppe, P.~D. Nguyen, and S.~Wermter, ``From semantics to execution:
  Integrating action planning with reinforcement learning for robotic tool
  use,'' \emph{arXiv preprint arXiv:1905.09683}, 2019.

\bibitem{paxton2019prospection}
C.~Paxton, Y.~Bisk, J.~Thomason, A.~Byravan, and D.~Fox, ``Prospection:
  Interpretable plans from language by predicting the future,'' in \emph{2019
  International Conference on Robotics and Automation (ICRA)}, 2019, pp.
  6942--6948.

\bibitem{bisk2016natural}
Y.~Bisk, D.~Yuret, and D.~Marcu, ``Natural language communication with
  robots,'' in \emph{Proceedings of the 2016 Conference of the North American
  Chapter of the Association for Computational Linguistics: Human Language
  Technologies}, 2016, pp. 751--761.

\bibitem{fu2019language}
J.~Fu, A.~Korattikara, S.~Levine, and S.~Guadarrama, ``From language to goals:
  Inverse reinforcement learning for vision-based instruction following,''
  \emph{ICLR}, 2019.

\bibitem{miller1995wordnet}
G.~A. Miller, ``{WordNet}: a lexical database for english,''
  \emph{Communications of the ACM}, vol.~38, no.~11, pp. 39--41, 1995.

\bibitem{pedersen2004wordnet}
T.~Pedersen, S.~Patwardhan, and J.~Michelizzi, ``Wordnet:: Similarity:
  measuring the relatedness of concepts,'' in \emph{Demonstration papers at
  HLT-NAACL 2004}.\hskip 1em plus 0.5em minus 0.4em\relax Association for
  Computational Linguistics, 2004, pp. 38--41.

\bibitem{artzi2013weakly}
Y.~Artzi and L.~Zettlemoyer, ``Weakly supervised learning of semantic parsers
  for mapping instructions to actions,'' \emph{Transactions of the Association
  for Computational Linguistics}, vol.~1, pp. 49--62, 2013.

\bibitem{williams2018learning}
E.~C. Williams, N.~Gopalan, M.~Rhee, and S.~Tellex, ``Learning to parse natural
  language to grounded reward functions with weak supervision,'' in \emph{2018
  IEEE International Conference on Robotics and Automation (ICRA)}.\hskip 1em
  plus 0.5em minus 0.4em\relax IEEE, 2018, pp. 1--7.

\bibitem{ross2018grounding}
C.~Ross, A.~Barbu, Y.~Berzak, B.~Myanganbayar, and B.~Katz, ``Grounding
  language acquisition by training semantic parsers using captioned videos,''
  in \emph{Proceedings of the 2018 Conference on Empirical Methods in Natural
  Language Processing}, 2018, pp. 2647--2656.

\end{thebibliography}

\end{document}